\documentclass[lettersize,journal]{IEEEtran}

\usepackage{amsmath,amsfonts}
\usepackage{array}
\usepackage[caption=false,font=normalsize,labelfont=sf,textfont=sf]{subfig}
\usepackage{textcomp}
\usepackage{stfloats}
\usepackage{url}
\usepackage{verbatim}
\usepackage{graphicx}
\usepackage{cite}

\usepackage{times}
\usepackage{epsfig}
\usepackage{epstopdf}
\usepackage{amssymb}
\usepackage{scalerel}
\usepackage{booktabs}
\usepackage{multirow}
\usepackage{adjustbox}
\usepackage{soul}
\usepackage[dvipsnames]{xcolor}

\usepackage{pifont}
\usepackage{xspace}
\usepackage{gensymb}
\usepackage{multirow}
\usepackage{algpseudocode}
\newcommand{\BF}[1]{\textbf{#1}}
\newcommand{\UL}[1]{\underline{#1}}
\newcommand{\shortname}{MAN\xspace}
\newcommand{\sxl}[1]{{\textcolor{black}{#1}}}
\newcommand{\xuelun}[1]{{\textcolor{black}{#1}}}
\newcommand{\parsection}[1]{\vspace{1mm}\noindent\textbf{#1:}~}

\begin{document}
\title{A Detector-oblivious Multi-arm Network for Keypoint Matching}
\author{
    Xuelun Shen,
    Qian Hu,~\IEEEmembership{Student Member,~IEEE,} \\
    Xin Li,
    Cheng Wang$^\ast$,~\IEEEmembership{Senior Member,~IEEE,}
    \thanks{Corresponding author: Cheng Wang.}
    \thanks{X. Shen, J. Zhang, Q. Hu, C. Wang are with Spatial Sensing and Computing Laboratory, School of Informatics, Xiamen University, China (e-mail: cwang@xmu.edu.cn).}
    \thanks{X. Li is with the School of Performance, Visualization, and Fine Arts, Texas A\&M University, USA.}
}
\maketitle

\begin{abstract}
    This paper presents a matching network to establish point correspondence between images. We propose a Multi-Arm Network (MAN) \xuelun{capable of learning} region overlap and depth, which can greatly improve keypoint matching robustness while bringing \sxl{an extra $50\%$ of computational time} during the inference stage. \xuelun{By adopting a} different \xuelun{design} from the state-of-the-art learning based pipeline SuperGlue \xuelun{framework, which} requires retraining when a different keypoint detector is adopted, our network can directly work with different keypoint detectors without time-consuming retraining processes. Comprehensive experiments conducted on four public benchmarks involving both outdoor and indoor \xuelun{scenarios demonstrate} that our proposed MAN outperforms state-of-the-art methods.
\end{abstract}

\section{Introduction}
\IEEEPARstart{F}{inding} keypoint correspondences is a fundamental computer vision problem.
Keypoint correspondences are often solved through matching local features under certain geometric constraints (e.g., mutual nearest and ratio tests).
Matching algorithms calculate similarities between the keypoint of an image and \emph{all} keypoints on another image.

\begin{figure}[!t]
\centering
\newcommand{\wid}{\columnwidth}
\subfloat[Correspondences by SuperGlue                           ]{\includegraphics[width=\wid]{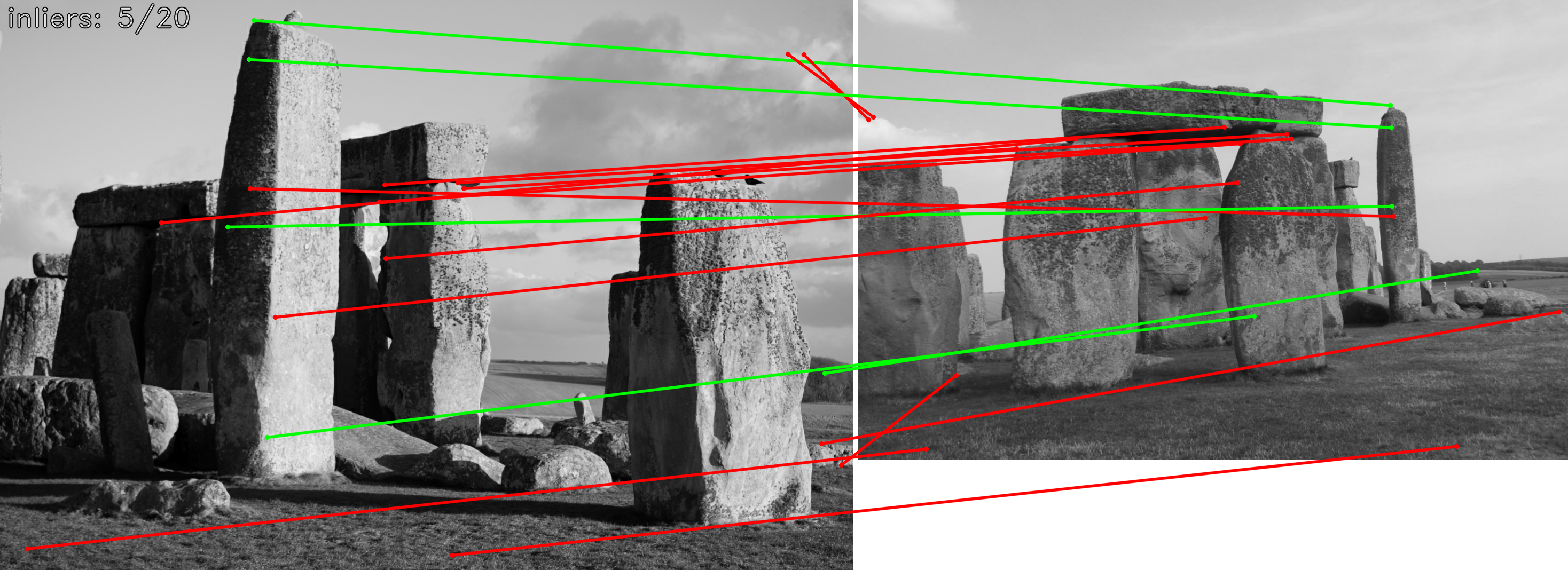}}\vspace{1mm}
\subfloat[Correspondences by \shortname                          ]{\includegraphics[width=\wid]{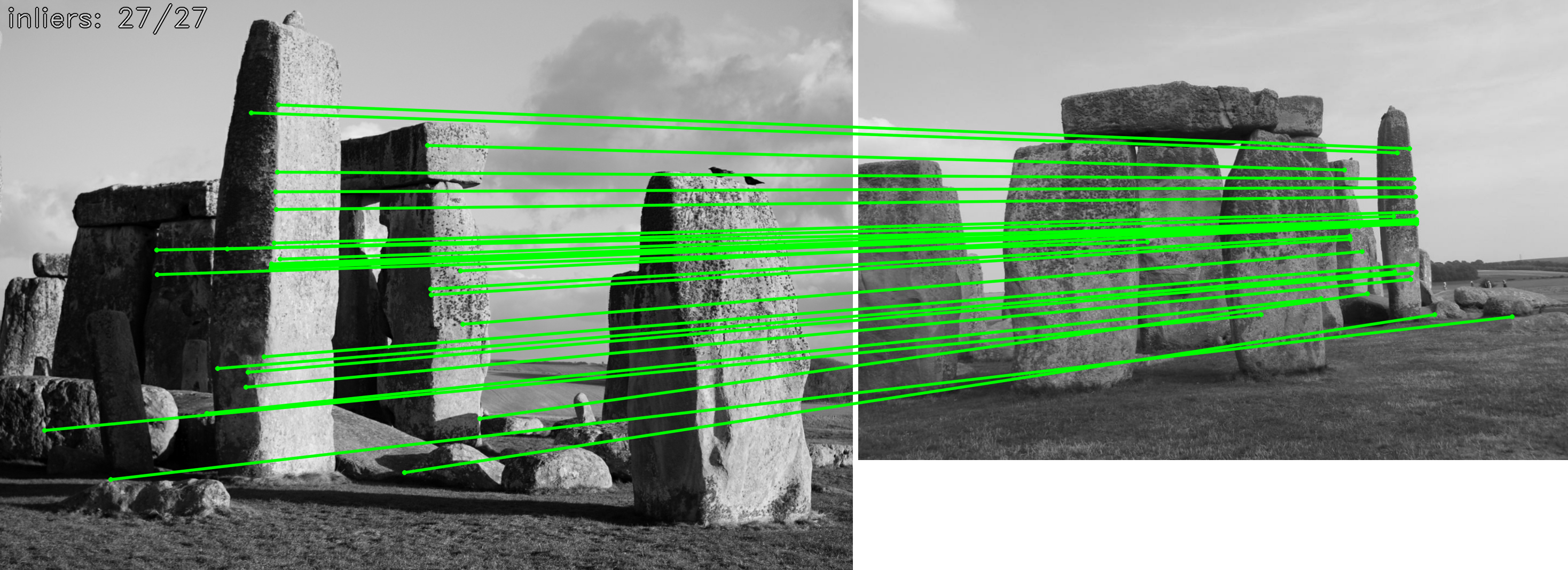}}
\caption{
    When images under drastic viewpoint and scale changes, the state-of-the-art pipeline SuperGlue~\cite{GLUE2020} fails to produce reliable matching.
    However our \shortname can more effectively find the correspondences.
    One example is illustrated here, where {\color{red}red} and {\color{green}green} lines indicate incorrect and correct matches, respectively.
}
\label{fig:Teaser}
\vspace{-\baselineskip}
\end{figure}

In this work, we expect the matching network to be aware of region-level information after communicating cross images.
We propose keypoint matching combined with solving two auxiliary tasks, through which regional-level correspondence \xuelun{can be} estimated.
Specifically, \xuelun{when} the network estimates a feature for each keypoint, we expect \xuelun{that} this feature could contain an overlap region and depth region information \xuelun{related to} the keypoint.
Fig.~\ref{fig:Teaser} \xuelun{shows} the reliable correspondences produced by our method \xuelun{and} designed \xuelun{using} the abovementioned idea, \xuelun{showing that this method is more effective} than a current state-of-the-art matching approach (SuperGlue~\cite{GLUE2020}).

\begin{table}
\centering
\renewcommand{\arraystretch}{1.3} 
\caption{
\sxl{
    Influence of keypoint \emph{coordinates} $\mathbf{p}$, \emph{descriptions} $\mathbf{d}$, and \emph{confidence} $\mathbf{c}$ for pose estimation in MegaDepth.
    $@X\degree$ represents the error thresholds,
    SP \xuelun{denotes that} the value is from a pretrained SuperPoint detector.
    RAND, ZERO, and ONE represent the value \xuelun{when the system} is set at random, all zeros, and all ones.
}
}
\vspace{-2mm}
\resizebox{0.7\linewidth}{!}{\begin{tabular}{ccc|lll}
    \toprule[1.5pt]
    \multirow{2}{*}{\sxl{p}}        & \multirow{2}{*}{\sxl{d}}    & \multirow{2}{*}{\sxl{c}}    & \multicolumn{3}{c}{\sxl{MegaDepth}}                            \\
    \cmidrule(lr){4-6}
                              &                       &                       & \sxl{@$5^\circ$}        & \sxl{@$10^\circ$}       & \sxl{@$20^\circ$}       \\
    \midrule
    \sxl{SP}                        & \sxl{SP}                    & \sxl{SP}                    & \sxl{$39.90$}           & \sxl{$52.49$}           & \sxl{$63.07$}           \\
    \sxl{SP}                        & \sxl{SP}                    & \sxl{RAND}                  & \sxl{$\mathbf{40.18}$}  & \sxl{$\mathbf{52.64}$}  & \sxl{$\mathbf{63.21}$}  \\
    \sxl{SP}                        & \sxl{SP}                    & \sxl{ZERO}                  & \sxl{$36.96$}           & \sxl{$49.88$}           & \sxl{$60.59$}           \\
    \sxl{SP}                        & \sxl{SP}                    & \sxl{ONE}                   & \sxl{$38.70$}           & \sxl{$51.59$}           & \sxl{$62.74$}           \\
    \bottomrule[1.5pt]

\end{tabular}}
\label{tab:teaser}
\vspace{-\baselineskip}
\end{table}

SuperGlue~\cite{GLUE2020} demonstrates remarkable performance in both outdoor and indoor datasets for keypoint matching, however, it \xuelun{requires} keypoint \emph{coordinates}, \emph{descriptions}, and \emph{confidence}; hence it needs to be retrained each time it adapts to different keypoint detectors.
This training process \xuelun{requires} a half month or more, even \xuelun{when using} the latest graphics cards (V100).
\sxl{
After conducting experiments on the influence of three keypoint \xuelun{attributes to} SuperGlue, we find that keypoint \emph{confidence} has little effect on performance.
Hence, we hypothesize that keypoint \emph{confidence} contributes little to the matching results (see Table~\ref{tab:teaser}).
}
This discovery provides experimental support for removing keypoint \emph{confidence}; this is an important step \xuelun{for} proposing detector-oblivious networks.
We introduce a new design into the matching pipeline, the detector-oblivious description network, which \xuelun{uses} keypoint \emph{coordinates} from any detectors \xuelun{as inputs} and outputs unique keypoint \emph{descriptions}.
This design \xuelun{entails that} our pipeline does not need to be retrained when a new detector is adopted.
In summary, our contributions are threefold:
\begin{itemize}
	\item
	We propose a Multi-Arm Network (MAN) for keypoint matching, which utilizes \xuelun{regional-level correspondence (overlap and depth)} to enhance the robustness of keypoint matching.
	\item
	We develop a detector-oblivious description network \xuelun{ensuring} that the matching pipeline is able to adapt to any keypoint detector without \xuelun{the requirements of} a time-consuming retraining procedure.
    \item
	Comprehensive experiments using four public benchmarks demonstrate that our pipeline \xuelun{facilitates} state-of-the-art results on both outdoor and indoor datasets.
\end{itemize}

\section{Related work}

Keypoints on images are usually described \xuelun{in terms of} their local features.
Generally, we estimate geometric transformations between images by matching local features.
This matching process can be conducted using learned networks.

\subsection{Matching keypoints using local features}
Matching keypoints using local features \xuelun{typically consists of} a nearest neighbor search \xuelun{coupled with} additional constraints.
These constraints \xuelun{can include} the mutual nearest neighbor, ratio tests~\cite{SIFT2004}, and heuristics (e.g., neighborhood consensus~\cite{GMS2017}).
Deep learning studies generally tend to obtain better \xuelun{and more distinctive} keypoint descriptors.
L2Net~\cite{L2NET2017} proposes learning discriminative descriptors in Euclidean space.
HardNet~\cite{HARD2017} introduces a loss function for learning local feature descriptors inspired by Lowe's matching criterion for SIFT~\cite{SIFT2004}.
He et al.~\cite{LDOA2018} improved the learning of local features by directly optimizing a ranking-based retrieval performance metric, \xuelun{termed} Average Precision.
LF-Net~\cite{LFNet2018} presents a deep architecture and training strategy \xuelun{by which} a local feature pipeline \xuelun{can be determined} from scratch.
RF-Net~\cite{RFNet2019} proposes an end-to-end trainable matching network based on the receptive field.
SOSNet~\cite{SOS2019} incorporates Second Order Similarity Regularization \xuelun{for learning} local feature descriptors.
Ebel et al.~\cite{POLAR2019} \xuelun{proposed that} the support region of local features \xuelun{may be determined by} directly \xuelun{using} a log-polar sampling scheme.
HyNet.~\cite{HYNET2020} introduced a hybrid similarity measure for triplet margin loss, a regularization term constraining the \xuelun{normalization of the} descriptor, and a new network architecture \xuelun{capable of performing} L2 normalization of all intermediate feature maps and their output descriptors.
Wang et al.~\cite{CPOSE2020} proposed a weakly-supervised framework \xuelun{capable of learning} feature descriptors solely from relative camera poses between images.
Shen et al.~\cite{LSA2022} explored a scale awareness learning approach for finding pixel-level correspondences based on the intuition that keypoints need to be extracted and described on appropriate scales.

\subsection{Matching keypoints using learned networks}
Matching keypoints using local features still find correspondences by nearest neighbor searches, discards the feature geometric information and ignores assignment structures.
SuperGlue~\cite{GLUE2020} proposes learning this information for matching.
It firstly merges keypoints coordinates, descriptions, and confidence by a hierarchical  Transformer~\cite{ATTE2017},
then uses the Sinkhorn algorithm~\cite{SINKD2013,SINK1967} to approximately solve the optimal transport problem~\cite{OPTI2008}.

\xuelun{Because} keypoint detectors are trained under different loss functions and designed by different algorithms, they generally give different descriptions (and confidence) to the same keypoint coordinate.
Therefore, \xuelun{it is necessary} to retrain SuperGlue~\cite{GLUE2020} each time it \xuelun{is combined} with different keypoint detectors,
\xuelun{because} the estimated match relies on keypoints attribution.
Nevertheless, it is \xuelun{possible} to \xuelun{use this} matching network \xuelun{using} the approach \xuelun{presented} in the Method section.




\subsection{Learning with auxiliary tasks}
Learning with auxiliary tasks to enhance the \xuelun{performance of the primary task} is not a new \xuelun{concept for} deep learning.
However, auxiliary tasks \xuelun{cannot be borrowed directly from} other networks without \xuelun{some consideration}.
Rich~\cite{MUTI1997} \xuelun{noted} that if auxiliary tasks are unrelated to the primary task,
the learned features in \xuelun{their} shared weights can be contaminated by outlier training signals from the unrelated auxiliary tasks; \xuelun{this can} result in performance degradation of the primary task.
Therefore, \xuelun{it is necessary} to find auxiliary tasks \xuelun{having} mutually beneficial relationships with the primary task.
Mordan et al.~\cite{ROCK2018} \xuelun{developed a means of estimating} the depth and normal to enhance the precision of object detection.
Zhou et al.~\cite{EGOM2017} \xuelun{enhanced} the pose estimation from depth prediction.
In this work, we estimate overlap and depth regions \xuelun{in order} to enhance the keypoint matching.

\begin{figure*}[t!]
\centering
\includegraphics[width=\linewidth]{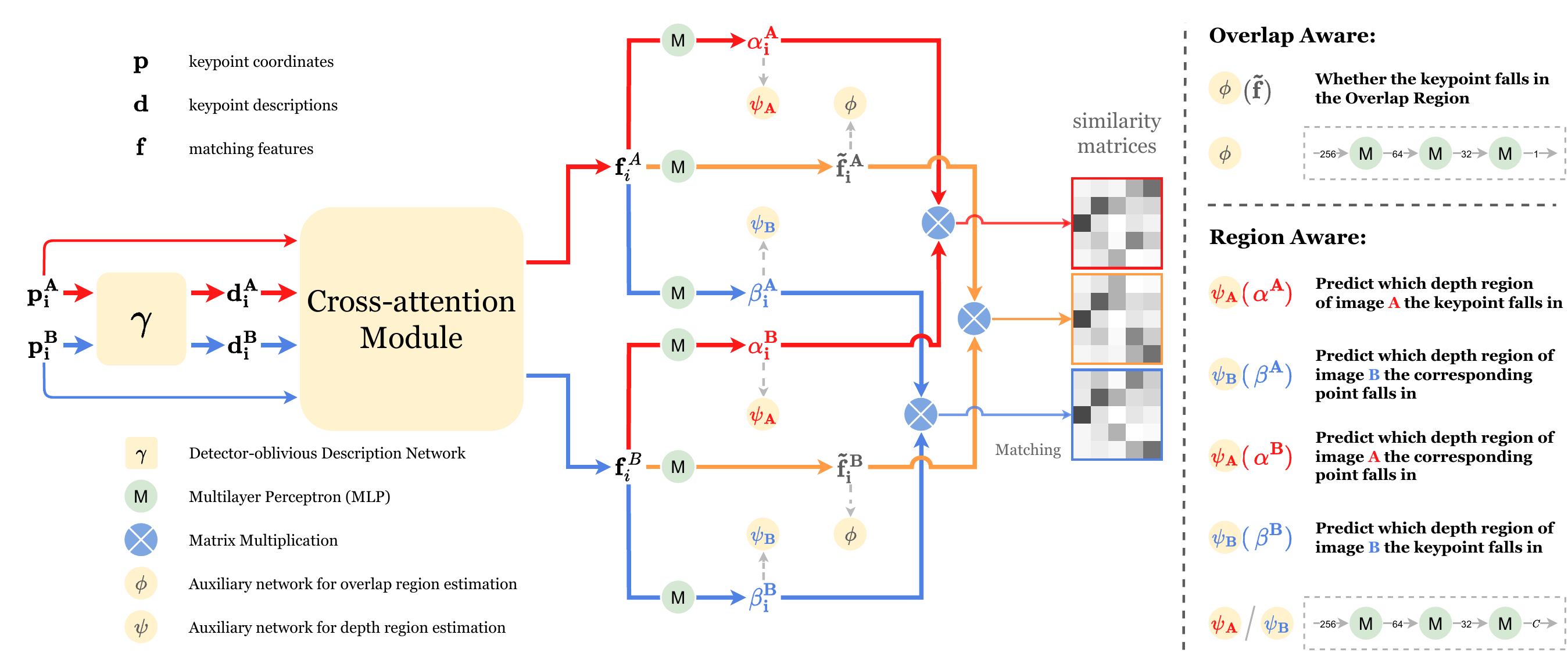}\vspace{0mm}
\caption{
	\xuelun{Architecture of the proposed} detector-oblivious multi-arm network (MAN).
	The functions $\mathop{\phi}$ and $\mathop{\psi}$ represent the auxiliary networks for overlap and depth regions estimation, respectively.
	Symbols in \textcolor{red}{red} and \textcolor{blue}{blue} denote the operations whose target images are images \textcolor{red}{A} and \textcolor{blue}{B}.
	}
\label{fig:Model}
\vspace{-\baselineskip}
\end{figure*}

\begin{figure}
\centering
\includegraphics[width=\linewidth]{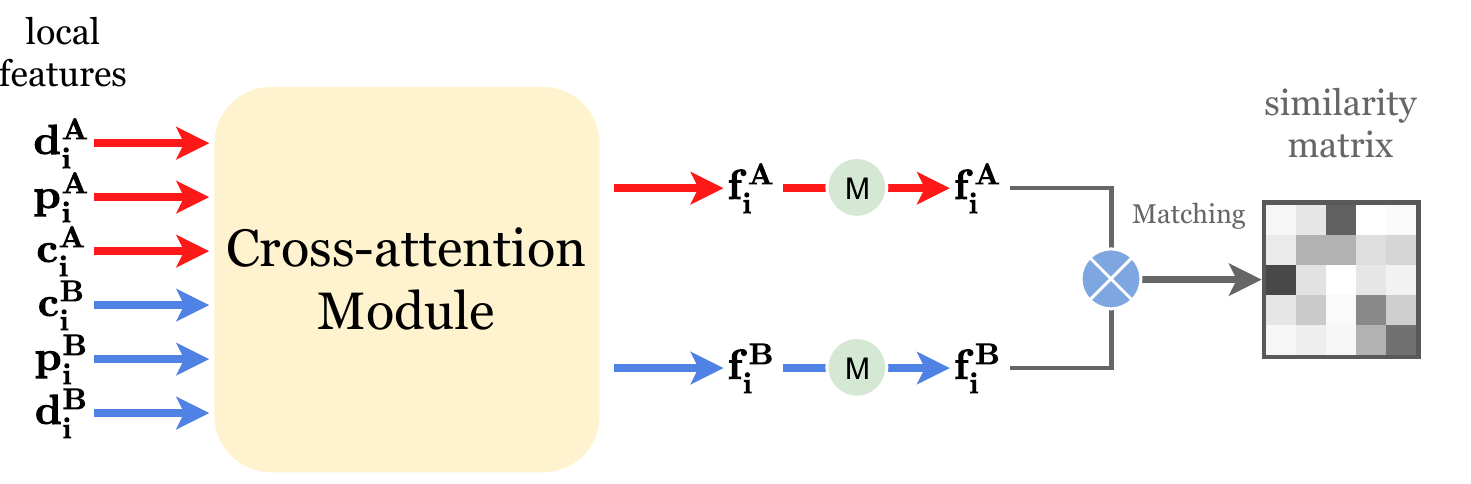}
\caption{
	Architecture of SuperGlue~\cite{GLUE2020},
	where $\mathbf{d}$, $\mathbf{p}$, and $\mathbf{c}$ represent keypoint \emph{descriptions}, \emph{coordinates}, and \emph{confidence}, respectively.}
\label{fig:SG}
\vspace{-\baselineskip}
\end{figure}

\section{Method}
In this section, we introduce our detector-oblivious MAN for keypoint matching.
Sec.~\ref{sec:PROBLEM} \xuelun{presents the formulation of} the problem \xuelun{of} keypoint matching.
Sec.~\ref{sec:FREE} introduces the approach \xuelun{by which} the detector-oblivious network \xuelun{is built}.
Sec.~\ref{sec:MAN} introduces the auxiliary tasks and loss functions developed in the MAN for overlap and depth region estimation.

\subsection{Problem formulation}
\label{sec:PROBLEM}

\xuelun{We consider} a pair of images $\mathbf{A}$ and $\mathbf{B}$, each with a set of keypoint \emph{coordinates} $\mathbf{p}$, which consist of $\mathbf{x}$ and $\mathbf{y}$ coordinates, $\mathbf{p}_{i}:=(\mathbf{x}, \mathbf{y})_{i}$.
Images $\mathbf{A}$ and $\mathbf{B}$ have $M$ and $N$ keypoint \emph{coordinates}, indexed by $\mathcal{A}:=\{1, \ldots, M\}$ and $\mathcal{B}:=\{1, \ldots, N\}$, respectively.
Our \xuelun{objective} is to train a network \xuelun{capable of outputting} a similarity matrix~\cite{GLUE2020} $\mathbf{M} \in[0,1]^{M \times N}$ from two sets of keypoint \emph{coordinates}.

\subsection{Detector-oblivious approach}
\label{sec:FREE}

As illustrated in Fig.~\ref{fig:SG}, SuperGlue requires input with keypoint \emph{descriptions} $\mathbf{d}$, \emph{coordinates} $\mathbf{p}$, and \emph{confidence} $\mathbf{c}$, which \xuelun{are represented} as \emph{local features}.
Descriptions $\mathbf{d}_{i} \in \mathbb{R}^{D}$ and confidence $\mathbf{c} \in [0, 1]$ can be extracted \xuelun{using} a network, e.g., SuperPoint~\cite{SPOINT2018} and R2D2~\cite{R2D22019}, or an algorithm, e.g., SIFT~\cite{SIFT2004}.

According to \xuelun{the} results \xuelun{of our experiments} in Table~\ref{tab:FREE-OUT-IN}, \xuelun{training} SuperGlue under SuperPoint~\cite{SPOINT2018}, \xuelun{followed by integration} with other detectors, e.g., R2D2~\cite{R2D22019}, \xuelun{results in poor} performance.
Because SuperPoint and R2D2 are trained under different network architectures and loss functions, they will \xuelun{result in} two completely different keypoint \emph{descriptions} $\mathbf{d}$ and \emph{confidence} $\mathbf{c}$ for the same keypoint.
SuperGlue can learn \xuelun{only} one detector through training and cannot transfer \xuelun{directly} to other detectors.

\xuelun{To determine} the impact of the three input parameters ($\mathbf{d}$, $\mathbf{p}$, and $\mathbf{c}$) on performance, we conducted a series of experiments (see Section~\ref{exp:free} \xuelun{for details}),
\xuelun{showing} that $\mathbf{p}$ and $\mathbf{d}$ are the \xuelun{primary contributing factors} to performance, \xuelun{whereas} $\mathbf{c}$ \xuelun{exerts a very minimal effect. Since} $\mathbf{c}$ contributes \xuelun{very little} to performance, \xuelun{we explored the impact of its removal.}

Furthermore, we propose a detector-oblivious description function $\mathop{\gamma}(\mathbf{I}, \mathbf{p})$ \xuelun{capable of obtaining} the keypoint \emph{descriptions} $\mathbf{d}$, \xuelun{which uses} an image $\mathbf{I} \in \{\mathbf{A}, \mathbf{B}\}$ and keypoint \emph{coordinates} $\mathbf{p}$ \xuelun{as the input}.
In practice, we implement function $\mathop{\gamma}$ \xuelun{using} a network \xuelun{capable of converting} keypoint \emph{descriptions} $\mathbf{d}$ from a required input parameter to the intermediate value of the network.
The network used to implement $\mathop{\gamma}$ could be any network \xuelun{satisfying} the formula
$\mathbf{d} = \mathop{\gamma}(\mathbf{I}, \mathbf{p})$.
If a network can provide dense feature maps for the input image $\mathbf{I}$, it can be used to implement $\mathop{\gamma}$.

\xuelun{Consequently}, we built a detector-oblivious network architecture \xuelun{as shown in} Fig.~\ref{fig:Model}.
This network is only required to be fed with the keypoint \emph{coordinates} $\mathbf{p}$ and can adapt to any keypoint detector \xuelun{following a single} training process.

\vspace{-3mm}
\subsection{The multi-arm network}
\label{sec:MAN}

Existing matching methods find correspondences directly at the pixel level.
Here, we attempt to enable the network to perceive correspondences at the region level (e.g., the overlap and depth regions).
The concept of the correspondence region depends both on images $\mathbf{A}$ and $\mathbf{B}$, rather than just \xuelun{a single} image.
Therefore, descriptors extracted from either image are insufficient to infer \xuelun{relationships between the regions}.
The keypoint descriptors $\mathbf{d}_{i}^{A}$ and $\mathbf{d}_{i}^{B}$ are required to communicate with each other for region estimation.
We borrowed the cross-attention module proposed by SuperGlue~\cite{GLUE2020} to communicate region-level information between images.
When the network fuses regional information into features, we use two auxiliary networks to estimate the overlap and depth regions.
In contrast with SuperGlue, which estimates only one similarity matrix to predict the matching result, our MAN grows three feature arms ($\mathbf{\tilde{f}}$, $\mathbf{\alpha}$, and $\mathbf{\beta}$ in~Fig.~\ref{fig:Model}) to estimate three similarity matrices.
For inference, we keep only the matching results that survive in all three similarity matrices, \xuelun{thereby increasing the reliability of} the matching (see Fig.~\ref{fig:Teaser}).

\parsection{Overlap Region Estimation Task}
As shown in Fig.~\ref{fig:Model}, our detector-oblivious network is fed with the \emph{keypoint coordinates} $\mathbf{p}$, and outputs the \emph{keypoint features} $\mathbf{d}_{i}$.
We use the cross-attention module to produce \emph{matching features} $\mathbf{f}$, then we use a Multilayer Perceptron (MLP) to transform $\mathbf{f}$ to a new high-dimensional space, represented as $\mathbf{\tilde{f}}$, which may be treated as new representations for the keypoint.
Before \xuelun{conducting} the matching process, we expect that features ~$\mathbf{\tilde{f}}$~ contain the information \xuelun{needed to determine} whether their keypoints fall into the overlap area.
To this end, we propose the OVerlap Estimation (OVE) Loss $\mathcal{L}_{\scaleto{\text{OVE}}{3.5pt}}$ to train the auxiliary network $\mathrm{\phi}$ :
\begin{equation}
	\mathbf{\upsilon}^{A}_{i}=
	\mathrm{\phi}
	\left(
		\mathbf{\tilde{f}}^{A}_{i}
	\right)=
	\mathrm{Sigmoid}
	\left(
	\mathrm{M L P}
	\left(
		\mathbf{\tilde{f}}^{A}_{i}
	\right)
	\right),
\end{equation}
\begin{equation}
	\mathcal{L}^{A}_{\scaleto{\text{OVE}}{3.5pt}} = \mathcal{L}_{\text {cross-entropy }}(y^{A}_{i}, \mathbf{\upsilon}^{A}_{i}),
\end{equation}
\begin{equation}
	\mathcal{L}_{\scaleto{\text{OVE}}{3.5pt}} = \mathcal{L}^{A}_{\scaleto{\text{OVE}}{3.5pt}} + \mathcal{L}^{B}_{\scaleto{\text{OVE}}{3.5pt}},
\end{equation}
where $\mathbf{\upsilon}^{A}_{i} \in (0, 1)$, and $y^{A}_{i} \in \{0, 1\}$ is the binary label.

\parsection{Depth Region Estimation Task}
The overlap region estimation task is \xuelun{required} to train features $\mathbf{\tilde{f}}$ to predict whether their keypoints fall into the overlap area.
The depth region estimation task trains features ($\mathbf{\alpha}$ and $\mathbf{\beta}$) \xuelun{in order} to predict into \emph{which depth region} of images $\mathbf{A}$ and $\mathbf{B}$ it falls.
We use Otsu's method~\cite{OTSU1979} to split the image into close-scene and far-scene \xuelun{regions according to} image depth.
Hence, the depth region estimation auxiliary network $\mathop{\psi}$ also \xuelun{performs} a binary classification.

\begin{figure}[t]
\centering
\newcommand{\wid}{0.96\columnwidth}
\includegraphics[width=\wid]{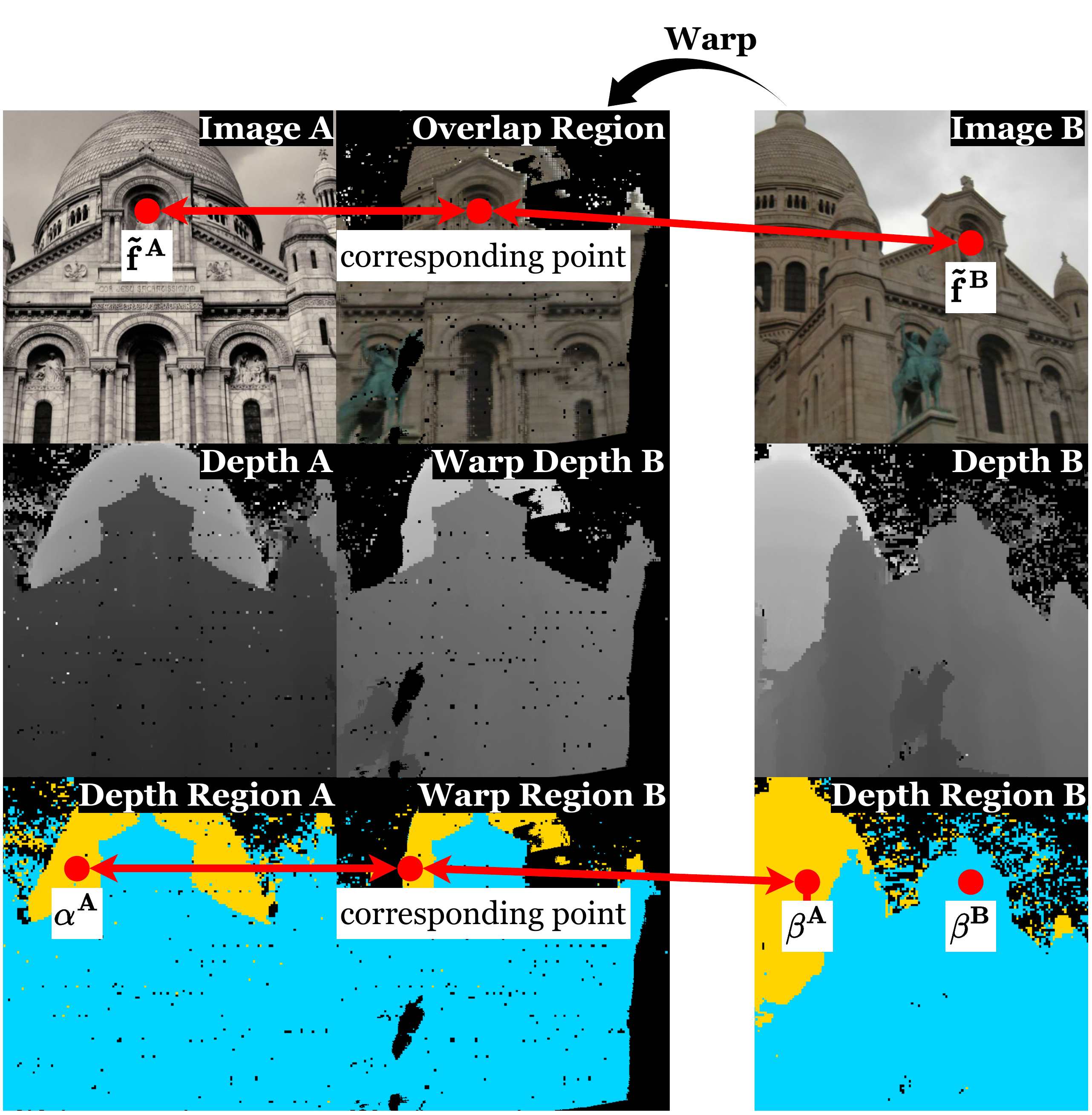}
\caption{
	Data visualization \xuelun{corresponding to} the depth region estimation task.
	\xuelun{Consider} a pair of images $\mathbf{A}$ and $\mathbf{B}$ and their corresponding depth maps $\mathbf{D}^{A}$ and $\mathbf{D}^{B}$.
	The depth invalid area is \xuelun{shown in} black.
	We divide the image into $C=3$ regions according to depth values, of which the valid regions occupy two regions (blue and yellow) and the invalid region occupies one (black).
}
\label{fig:Depth}
\end{figure}

Fig.~\ref{fig:Depth} shows the data for training network $\mathop{\psi}$ given a pair of images ($\mathbf{A}$ and $\mathbf{B}$) and their corresponding depth maps ($\mathbf{D}^{A}$ and $\mathbf{D}^{B}$).
We divide images into $C=3$ regions according to depth values, of which the valid regions occupy two regions (blue and yellow) and the invalid region occupies one (black), as follows
\begin{equation}
	\epsilon =
	\mathrm{OTSU}
	\left(
		\mathbf{D}
	\right)
\end{equation}
\begin{equation}
	c =
	\left\{\begin{array}
		{ll}
		0 & \text { if } \; \mathbf{D}_{ij} \in [\mathbf{D}^{min}, \epsilon ), \\
		1 & \text { if } \; \mathbf{D}_{ij} \in [\epsilon, \mathbf{D}^{max} ], \\
		2 & \text { if } \; \mathbf{D}_{ij} \; is \; \text {invalid},
	\end{array}\right.
\end{equation}
where
$\mathbf{D}$ represents the depth map of an image,
$\mathbf{D}^{min}$ and $\mathbf{D}^{max}$ represent the minimum and maximum values of a depth map,
and $c$ represents the region class.

As illustrated in Fig.~\ref{fig:Model}, our detector-oblivious network estimates a feature $\mathbf{f^{A}_{i}}$ to describe a keypoint on the coordinate $\mathbf{p^{A}_{i}}$ in image A.
Then, two MLPs \xuelun{are used to} obtain features
\footnote{For better understanding, we color symbols in \textcolor{red}{red} and \textcolor{blue}{blue} for the operations whose target images are images \textcolor{red}{A} and \textcolor{blue}{B}.}
\textcolor{red}{$\mathbf{\alpha^{A}_{i}}$} and
\textcolor{blue}{$\mathbf{\beta^{A}_{i}}$}.
For these two features, we \xuelun{anticipate the following}
\begin{itemize}
	\item \textcolor{red}{$\mathbf{\alpha^{A}_{i}}$} can be used to estimate which region of image \textcolor{red}{$\mathbf{A}$} the keypoint $\mathbf{p^{A}_{i}}$ falls in,
	\item \textcolor{blue}{$\mathbf{\beta^{A}_{i}}$} can be used to estimate which region of image \textcolor{blue}{$\mathbf{B}$} the keypoint $\mathbf{p^{A}_{i}}$'s corresponding point falls in.
\end{itemize}
To this end, we propose the DEpth region Estimation (DEE) Loss \textcolor{red}{$\hat{\mathcal{L}}^{A}_{\scaleto{\text{DEE}}{3.5pt}}$} as follows to train the auxiliary network \textcolor{red}{$\mathop{\psi_{\scaleto{A}{3.5pt}}}$} :
\begin{equation}
	\hat{\mathbf{\eta}}^{A}_{i}=\textcolor{red}{\mathop{\psi_{\scaleto{A}{3.5pt}}}}\left(\textcolor{red}{\mathbf{\alpha^{A}_{i}}}\right) = \mathrm{Sigmoid}\left(\mathrm{M L P\left(\textcolor{red}{\mathbf{\alpha^{A}_{i}}}\right)}\right),
\end{equation}
\begin{equation}
	\textcolor{red}{\hat{\mathcal{L}}^{A}_{\scaleto{\text{DEE}}{3.5pt}}} = \mathcal{L}_{\text {cross-entropy }}(\hat{y}^{A}_{i}, \hat{\mathbf{\eta}}^{A}_{i}),
\end{equation}
where $\hat{\mathbf{\eta}}^{A}_{i} \in (0, 1)$, and $\hat{y}^{A}_{i} \in \{0, 1\}$ are the binary labels.
Similarly, we obtain Loss \textcolor{blue}{$\bar{\mathcal{L}}^{A}_{\scaleto{\text{DEE}}{3.5pt}}$} from \textcolor{blue}{$\mathbf{\beta^{A}_{i}}$} to train network \textcolor{blue}{$\mathop{\psi_{\scaleto{B}{3.5pt}}}$}.

Feature $f^{A}_{i}$ grows two features \textcolor{red}{$\mathbf{\alpha^{A}_{i}}$} and \textcolor{blue}{$\mathbf{\beta^{A}_{i}}$}, which we expect to predict the regions of their own image \textcolor{red}{$\mathbf{A}$} and the corresponding image \textcolor{blue}{$\mathbf{B}$}.
\xuelun{The process} is similar for \textcolor{red}{$\mathbf{\alpha^{B}_{i}}$} and \textcolor{blue}{$\mathbf{\beta^{B}_{i}}$}.
For features that estimate \sxl{regions in the same target image}, we expect the following:
\begin{itemize}
	\item \textcolor{red}{$\mathbf{\alpha^{A}_{i}}$} to match \textcolor{red}{$\mathbf{\alpha^{B}_{i}}$} (their target regions are on image \textcolor{red}{$\mathbf{A}$}),
	\item \textcolor{blue}{$\mathbf{\beta^{A}_{i}}$} to match \textcolor{blue}{$\mathbf{\beta^{B}_{i}}$} (their target regions are on image \textcolor{blue}{$\mathbf{B}$}).
\end{itemize}

In summary, the final loss function $\mathcal{L}_{\scaleto{\text{DEE}}{3.5pt}}$ for the depth region estimation task can be formulated as follows:
\begin{equation}
	\mathcal{L}_{\scaleto{\text{DEE}}{3.5pt}} =
	\textcolor{red}{\hat{\mathcal{L}}^{A}_{\scaleto{\text{DEE}}{3.5pt}}}+
	\textcolor{red}{\hat{\mathcal{L}}^{B}_{\scaleto{\text{DEE}}{3.5pt}}}+
	\textcolor{blue}{\bar{\mathcal{L}}^{A}_{\scaleto{\text{DEE}}{3.5pt}}}+
	\textcolor{blue}{\bar{\mathcal{L}}^{B}_{\scaleto{\text{DEE}}{3.5pt}}}.
\end{equation}
\sxl{Since the keypoint that falls in the non-overlap area or invalid depth area cannot find the corresponding point in another image, the training data during the training, for $\mathcal{L}_{\scaleto{\text{DEE}}{3.5pt}}$ are composed only of keypoints that fall in the overlap and valid depth areas between the two images.}

\parsection{Multi-arm Network}
\xuelun{During} training, we added two auxiliary tasks (OVE and DEE) as mentioned above to make the network aware of the overlap and depth region; therefore, our total objective function is expressed as \xuelun{follows}:
\begin{equation}
	\mathcal{L}_{\scaleto{\text{summary}}{3.5pt}} =
	\mathcal{L}_{\scaleto{\text{OVE}}{3.5pt}}+
	\mathcal{L}_{\scaleto{\text{DEE}}{3.5pt}}+
	\mathcal{L}_{\scaleto{\text{Match}}{5pt}},
\end{equation}
where the Match Loss $\mathcal{L}_{\scaleto{\text{Match}}{5pt}}$ is proposed by SuperGlue \xuelun{ and minimizes} the negative log-likelihood of the similarity matrix $\mathbf{M}$.

For inference, the auxiliary networks ($\phi$ and $\psi$) can be detached, while its features ($\mathbf{\tilde{f}}$, $\mathbf{\alpha}$, and $\mathbf{\beta}$) grow on the three arms are reserved.
We infer the matching results from three similarity matrices and keep the matching results that survive in all three similarity matrices.

\begin{table*}
\centering
\renewcommand{\arraystretch}{1.3} 
\caption{
\sxl{
    Influence of keypoint coordinates $\mathbf{p}$, descriptions $\mathbf{d}$, and confidence $\mathbf{c}$ for outdoor and indoor pose estimation.
    The first column represents the number of the experiment.
}
}
\vspace{-2mm}
\resizebox{0.77\linewidth}{!}{\begin{tabular}{c|ccc|lllll|lllll}
    \toprule[1.5pt]
    \multirow{2}{*}{\# Outdoor}     & \multirow{2}{*}{p}        & \multirow{2}{*}{d}    & \multirow{2}{*}{c}    & \multicolumn{5}{c|}{MegaDepth}                                                                    &\multicolumn{5}{c}{YFCC}                                                                           \\
    \cmidrule(lr){5-9}
    \cmidrule(lr){10-14}
                                    &                           &                       &                       & @$5^\circ$        & @$10^\circ$       & @$20^\circ$       & P                 & M                 & @$5^\circ$        & @$10^\circ$       & @$20^\circ$       & P                 & M                 \\
    \midrule
    1                               & SP                        & SP                    & SP                    & $39.90$           & $52.49$           & $63.07$           & $\mathbf{92.88}$  & $25.91$           & $38.92$           & $59.12$           & $75.45$           & $\mathbf{98.72}$  & $23.57$           \\
    2                               & R2D2                      & SP                    & SP                    & $20.90$           & $30.98$           & $40.82$           & $87.09$           & $\mathbf{28.21}$  & $34.94$           & $54.97$           & $71.31$           & $97.43$           & $\mathbf{29.65}$  \\
    3                               & SP                        & R2D2                  & SP                    & $0.00$            & $0.14$            & $0.45$            & $1.01$            & $0.52$            & $0.00$            & $0.00$            & $0.00$            & $0.11$            & $0.03$            \\
    4                               & SP                        & SP                    & R2D2                  & $37.02$           & $49.68$           & $60.30$           & $92.35$           & $22.89$           & $38.83$           & $\mathbf{59.43}$  & $\mathbf{75.78}$  & $98.53$           & $23.28$           \\
    5                               & SP                        & SP                    & RAND                  & $\mathbf{40.18}$  & $\mathbf{52.64}$  & $\mathbf{63.21}$  & $92.76$           & $25.95$           & $\mathbf{39.10}$  & $59.29$           & $75.56$           & $98.57$           & $23.87$           \\
    6                               & SP                        & SP                    & ZERO                  & $36.96$           & $49.88$           & $60.59$           & $91.05$           & $25.95$           & $37.86$           & $58.27$           & $75.00$           & $98.48$           & $24.07$           \\
    7                               & SP                        & SP                    & ONE                   & $38.70$           & $51.59$           & $62.74$           & $92.81$           & $26.49$           & $38.60$           & $58.91$           & $75.46$           & $98.42$           & $24.44$           \\

    \toprule[1.5pt]
    \multirow{2}{*}{\# Indoor}      & \multirow{2}{*}{p}        & \multirow{2}{*}{d}    & \multirow{2}{*}{c}    & \multicolumn{5}{c|}{ScanNet}                                                                      &\multicolumn{5}{c}{SUN3D}                                                                          \\
    \cmidrule(lr){5-9}
    \cmidrule(lr){10-14}
                                    &                           &                       &                       & @$5^\circ$        & @$10^\circ$       & @$20^\circ$       & P                 & M                 & @$5^\circ$        & @$10^\circ$       & @$20^\circ$       & P                 & M                 \\
    \midrule
    1                               & SP                        & SP                    & SP                    & $\mathbf{16.45}$  & $\mathbf{33.62}$  & $\mathbf{51.96}$  & $\mathbf{84.13}$  & $31.68$           & $7.08$            & $18.05$           & $33.88$           & $87.07$           & $44.96$           \\
    2                               & R2D2                      & SP                    & SP                    & $4.66$            & $12.81$           & $24.00$           & $68.53$           & $16.33$           & $3.87$            & $11.13$           & $23.74$           & $83.23$           & $34.37$           \\
    3                               & SP                        & R2D2                  & SP                    & $0.00$            & $0.00$            & $0.00$            & $0.60$            & $0.01$            & $0.00$            & $0.00$            & $0.02$            & $0.83$            & $0.02$            \\
    4                               & SP                        & SP                    & R2D2                  & $16.04$           & $33.35$           & $51.69$           & $83.83$           & $\mathbf{31.87}$  & $7.07$            & $17.96$           & $33.86$           & $86.94$           & $45.04$           \\
    5                               & SP                        & SP                    & RAND                  & $15.54$           & $32.77$           & $50.47$           & $83.31$           & $31.78$           & $\mathbf{7.17}$   & $\mathbf{18.24}$  & $\mathbf{34.10}$  & $86.78$           & $\mathbf{45.20}$  \\
    6                               & SP                        & SP                    & ZERO                  & $15.23$           & $31.33$           & $49.23$           & $84.07$           & $31.41$           & $6.89$            & $17.54$           & $33.31$           & $\mathbf{87.11}$  & $44.76$           \\
    7                               & SP                        & SP                    & ONE                   & $14.70$           & $31.70$           & $49.34$           & $82.58$           & $31.61$           & $7.07$            & $18.14$           & $33.94$           & $86.32$           & $45.09$           \\
    \bottomrule[1.5pt]

\end{tabular}}
\label{tab:FREE-OUT-IN}
\vspace{-\baselineskip}
\end{table*}

\section{Experiments for the detector-oblivious approach}
\label{exp:free}

\sxl{In this section, we test} the influence of keypoint \emph{coordinates} $\mathbf{p}$, \emph{descriptions} $\mathbf{d}$, and \emph{confidence} $\mathbf{c}$ for keypoint matching \xuelun{using} SuperGlue~\cite{GLUE2020} in outdoor and indoor scenes.

\parsection{Metrics}
\xuelun{Following} previous work SuperGlue~\cite{GLUE2020} to report the area under the cumulative error curve (AUC) of the pose error at \xuelun{certain} thresholds ($5^\circ$,$10^\circ$, and $20^\circ$).
The pose error is the maximum of the angular errors in rotation and translation.
Poses are computed by estimating the essential matrix with OpenCV’s \emph{findEssentialMat} and RANSAC~\cite{RANSAC1981,RAGU2008} using an inlier threshold of one pixel divided by the focal length, followed by \emph{recoverPose}.
In addition, we also report the matching precision ($\mathbf{P}$) and the matching score ($\mathbf{M}$)~\cite{LIFT2016}, where a match is considered correct based on its epipolar distance.
The epipolar correctness threshold is $5\cdot10^{-4}$.
\sxl{
The matching precision is the ratio of the correct correspondences in the \xuelun{overall} correspondences \xuelun{found} by the method.
The matching score is the ratio of ground truth correspondences that can be recovered by the method over its extracted keypoint features.
}

\parsection{Baselines}
The original SuperGlue is trained under SuperPoint (SP) \xuelun{and recorded as} \textbf{\#1} in Table~\ref{tab:FREE-OUT-IN}.
To test the generalization of SuperGlue integrated with different detectors, we changed $\mathbf{p}$, $\mathbf{d}$, and $\mathbf{c}$ from the original SP to R2D2.

\parsection{Implement details}
The SuperGlue model used in this experiment is the official release model, which \xuelun{requires inputs in the form of} keypoint \emph{coordinates}($2$-dim), \emph{descriptions}($256$-dim), and \emph{confidence}($scalar$).
To fit the \emph{description} input dimension of $256$, we trained a $256$-dim R2D2~\cite{R2D22019} by its official release code (the original R2D2 output $128$-dim descriptions for the keypoints).

\begin{table*}
\centering
\renewcommand{\arraystretch}{1.3} 
\vspace{-2mm}\caption{
\sxl{
    Evaluation for outdoor and indoor pose estimation.
    We report the AUC of the pose error at the thresholds  ($5^\circ$,$10^\circ$, and $20^\circ$), the matching precision ($\mathbf{P}$), and the matching score ($\mathbf{M}$).
}
}
\vspace{-2mm}
\resizebox{0.66\linewidth}{!}{\begin{tabular}{l|l|rrrrr|rrrrr}
    \toprule[1.5pt]
    \multirow{2}{*}{\textbf{Outdoor}} & \multirow{2}{*}{Matcher}  & \multicolumn{5}{c|}{MegaDepth}                                                                     &\multicolumn{5}{c}{YFCC}                                                                           \\
    \cmidrule(l){3-7}
    \cmidrule(l){8-12}
    &
                                    & @$5^\circ$        & @$10^\circ$       & @$20^\circ$       & P                 & M                                     & @$5^\circ$        & @$10^\circ$       & @$20^\circ$       & P                 & M                 \\
    \midrule
    \multirow{6}{*}{SIFT}
    & NN + ratio test               & 17.00             & 24.68             & 32.67             & 60.87             & \multicolumn{1}{r|}{6.44}             & 22.51             & 35.27             & 48.52             & 57.13             & 3.50              \\
    & NN + PointCN                  & 12.70             & 18.57             & 25.62             & 54.41             & \multicolumn{1}{r|}{11.15}            & 23.05             & 37.03             & 51.06             & 57.57             & 10.33             \\
    & NN + OANet                    & 9.06              & 13.80             & 20.04             & 51.80             & \multicolumn{1}{r|}{6.83}             & 18.93             & 30.32             & 42.18             & 58.25             & 6.54              \\
    & NN + ACNe                     & 15.76             & 22.09             & 29.60             & 57.21             & \multicolumn{1}{r|}{14.49}            & 26.05             & 40.90             & 55.19             & 63.12             & 12.17             \\
    & SuperGlue                     & 24.77             & 35.19             & 45.82             & 89.66             & \multicolumn{1}{r|}{15.53}            & 37.28             & 55.47             & 70.65             & 95.60             & 15.37             \\
    & Multi-Arm                     & \BF{26.01}        & \BF{36.25}        & \BF{50.63}        & \BF{94.78}        & \multicolumn{1}{r|}{\BF{16.65}}       & \BF{38.44}        & \BF{56.94}        & \BF{72.25}        & \BF{96.34}        & \BF{16.59}        \\
    \midrule
    \multirow{6}{*}{SuperPoint}
    & NN + mutal                    & 17.19             & 25.51             & 35.25             & 42.68             & \multicolumn{1}{r|}{17.17}            & 16.68             & 30.18             & 45.84             & 39.36             & 12.23             \\
    & NN + PointCN                  & 21.88             & 31.56             & 41.20             & 66.28             & \multicolumn{1}{r|}{16.85}            & 23.26             & 40.00             & 57.08             & 74.61             & 17.85             \\
    & NN + OANet                    & 15.38             & 24.12             & 33.14             & 65.19             & \multicolumn{1}{r|}{10.65}            & 20.00             & 35.66             & 51.79             & 78.29             & 11.87             \\
    & NN + ACNe                     & 24.09             & 33.70             & 43.10             & 69.53             & \multicolumn{1}{r|}{21.92}            & 26.22             & 44.96             & 62.68             & 80.85             & 20.31             \\
    & SuperGlue                     & 39.90             & 52.49             & 63.07             & 92.88             & \multicolumn{1}{r|}{25.91}            & 38.92             & 59.12             & 75.45             & 98.72             & 23.57             \\
    & Multi-Arm                     & \BF{42.55}        & \BF{54.65}        & \BF{64.56}        & \BF{94.54}        & \multicolumn{1}{r|}{\BF{27.23}}       & \BF{42.65}        & \BF{62.61}        & \BF{77.87}        & \BF{99.82}        & \BF{24.64}        \\

    \midrule
    \multirow{2}{*}{\sxl{Detector-Free}}
    & \sxl{DFM}                     		& \sxl{30.22}             & \sxl{37.11}             & \sxl{45.44}             & \sxl{78.02}             & \multicolumn{1}{r|}{\sxl{22.41}}            & \sxl{30.51}             & \sxl{46.06}             & \sxl{64.17}             & \sxl{83.52}             & \sxl{21.03}             \\
    & \sxl{LoFTR}                     	& \BF{\sxl{44.94}}        & \BF{\sxl{58.29}}        & \BF{\sxl{70.80}}        & \BF{\sxl{98.46}}        & \multicolumn{1}{r|}{\BF{\sxl{29.03}}}       & \BF{\sxl{43.47}}        & \BF{\sxl{64.97}}        & \BF{\sxl{82.34}}        & \BF{\sxl{98.90}}        & \BF{\sxl{25.83}}        \\

    \toprule[1.5pt]
    \multirow{2}{*}{\textbf{Indoor}} & \multirow{2}{*}{Matcher}  & \multicolumn{5}{c|}{ScanNet}                                                                       &\multicolumn{5}{c}{SUN3D}                                                                          \\
    \cmidrule(l){3-7}
    \cmidrule(l){8-12}
    &
                                    & @$5^\circ$        & @$10^\circ$       & @$20^\circ$       & P                 & M                                     & @$5^\circ$        & @$10^\circ$       & @$20^\circ$       & P                 & M                 \\
    \midrule
    \multirow{6}{*}{SIFT}
    & NN + ratio test               & 5.38              & 11.83             & 21.20             & 46.15             & \multicolumn{1}{r|}{7.05}             & 4.80              & 12.28             & 24.21             & 67.51             & 14.26             \\
    & NN + PointCN                  & 3.22              & 9.09              & 17.95             & 66.31             & \multicolumn{1}{r|}{6.61}             & 4.49              & 11.87             & 23.63             & 73.73             & 18.86             \\
    & NN + OANet                    & 2.95              & 7.22              & 13.19             & 41.13             & \multicolumn{1}{r|}{6.25}             & 3.59              & 9.57              & 19.44             & 68.43             & 13.09             \\
    & NN + ACNe                     & 5.54              & 12.96             & 22.51             & 50.55             & \multicolumn{1}{r|}{12.33}            & 4.74              & 12.35             & 24.36             & 72.08             & 21.72             \\
    & SuperGlue                     & 7.80              & 19.50             & 33.88             & 76.95             & \multicolumn{1}{r|}{\BF{19.05}}       & 5.78              & 14.97             & 29.13             & 85.83             & \BF{32.40}        \\
    & Multi-Arm                     & \BF{10.00}        & \BF{22.03}        & \BF{35.75}        & \BF{79.94}        & \multicolumn{1}{r|}{17.53}            & \BF{5.95}         & \BF{15.85}        & \BF{29.62}        & \BF{86.77}        & 29.32             \\
    \midrule
    \multirow{6}{*}{SuperPoint}
    & NN + mutal                    & 9.45              & 22.46             & 37.41             & 51.85             & \multicolumn{1}{r|}{22.48}            & 6.28              & 16.02             & 30.78             & 70.42             & 37.02             \\
    & NN + PointCN                  & 3.59              & 9.77              & 20.53             & 77.31             & \multicolumn{1}{r|}{9.73}             & 4.34              & 12.25             & 25.70             & 84.37             & 29.83             \\
    & NN + OANet                    & 3.27              & 9.11              & 18.22             & 61.59             & \multicolumn{1}{r|}{9.80}             & 3.06              & 8.96              & 19.55             & 81.34             & 16.54             \\
    & NN + ACNe                     & 8.79              & 21.56             & 36.40             & 68.32             & \multicolumn{1}{r|}{26.52}            & 5.66              & 14.89             & 29.37             & 83.18             & 43.56             \\
    & SuperGlue                     & 16.45             & 33.62             & 51.96             & 84.13             & \multicolumn{1}{r|}{\BF{31.68}}       & 7.08              & 18.05             & 33.88             & 87.07             & \BF{44.96}        \\
    & Multi-Arm                     & \BF{17.65}        & \BF{35.09}        & \BF{55.61}        & \BF{84.73}        & \multicolumn{1}{r|}{29.11}            & \BF{7.24}         & \BF{18.65}        & \BF{35.38}        & \BF{90.70}        & 41.72             \\

    \midrule
    \multirow{2}{*}{\sxl{Detector-Free}}
    & \sxl{DFM}                     		& \sxl{11.75}             & \sxl{28.28}             & \sxl{42.97}             & \sxl{71.44}             & \multicolumn{1}{r|}{\sxl{22.18}}            & \sxl{5.72}             & \sxl{15.74}             & \sxl{30.87}             & \sxl{84.03}             & \sxl{42.60}             \\
    & \sxl{LoFTR}                     	& \BF{\sxl{17.73}}        & \BF{\sxl{35.61}}        & \BF{\sxl{57.89}}        & \BF{\sxl{88.44}}        & \multicolumn{1}{r|}{\BF{\sxl{32.48}}}       & \BF{\sxl{7.59}}        & \BF{\sxl{20.12}}        & \BF{\sxl{37.79}}        & \BF{\sxl{93.74}}        & \BF{\sxl{48.72}}        \\

    \bottomrule[1.5pt]
\end{tabular}}
\label{tab:MAN-OUT-IN}
\vspace{-\baselineskip}
\end{table*}

\vspace{-3mm}
\subsection{Outdoor scenarios}
\label{exp:free-out}

\parsection{Datasets}
In the outdoor scenario, we used the MegaDepth~\cite{MEGA2018} and YFCC100M datasets~\cite{YFCC2016} for testing.
MegaDepth consists of $196$ different scenes reconstructed from $1070468$ internet photos using COLMAP
\cite{SFM2016,PWISE2016}.
We selected eight challenge scenes (868 image pairs) for evaluation, \xuelun{similar to} the test data used for SuperGlue.
YFCC100M contains several touristic landmark images whose ground-truth poses were created by generating 3D reconstructions from a subset of the collections~\cite{RECON2015}.
We selected four scenes (4000 image pairs) for evaluation.
For testing, images were resized such that their longest dimensions were equal to $1600$ pixels.
We detected $2048$ keypoints in both R2D2 and SP.

\parsection{Results}
As shown in Table~\ref{tab:FREE-OUT-IN}, after changing the keypoint \emph{coordinates} $\mathbf{p}$ \xuelun{as shown in} record \textbf{\#2}, the performance of pose estimation is decreased \xuelun{by more} than 18\% \xuelun{relative to} MegaDepth.
In YFCC, \xuelun{although} this phenomenon is alleviated, the performance is still degraded.
After changing the keypoint \emph{descriptors} $\mathbf{d}$ \xuelun{as shown in} record \textbf{\#3}, the model did not \xuelun{work for either datasets}.
Because SuperGlue is only trained under the SP description, it cannot adapt to R2D2.
After replacing keypoint \emph{confidence} $\mathbf{c}$ from SP to R2D2, random numbers, zeros, and ones, \xuelun{as shown in records} \textbf{\#4} - \textbf{\#7}, in two datasets, the performance variance is small.
In addition, we obtained the best pose estimation result for MegaDepth when setting the keypoint \emph{confidence} randomly, rather than using the original SP.

\subsection{Indoor scenarios}
\label{exp:free-in}

\parsection{Datasets}
For the indoor scenario, we used the ScanNet~\cite{SCAN2017} and the SUN3D~\cite{SUN2013} datasets.
ScanNet is a large-scale indoor dataset composed of monocular sequences with ground truth poses and depth images.
We selected $1500$ image pairs from these well-defined test sequences for evaluation, which is \xuelun{similar to} the test data used in SuperGlue.
The SUN3D dataset comprises a series of indoor videos captured using a Kinect, with 3D reconstructions; we used $14872$ image pairs from the well-defined test sequences for evaluation.
For testing, images were resized such that their longest dimensions are equal to $640$ pixels.
We detected $1024$ keypoints for both R2D2 and SP.

\parsection{Results}
As shown in Table~\ref{tab:FREE-OUT-IN}, changing the keypoint \emph{coordinates} $\mathbf{p}$ \xuelun{as shown in} record \textbf{\#2} cause the pose estimation performance to degrade significantly in both ScanNet and SUN3D.
After changing keypoint \emph{descriptors} $\mathbf{d}$ \xuelun{as shown in} record \textbf{\#3}, in both ScanNet and SUN3D, the model does not work.
This result is consistent with experimental results in the outdoor scenario.
After replacing keypoint \emph{confidence} $\mathbf{c}$ from SP to R2D2, random numbers, zeros, and ones, \xuelun{as shown in records} \textbf{\#4} - \textbf{\#7}, both ScanNet and SUN3D show little change in performance.
In addition, we obtain the best pose estimation result in the SUN3D when setting the keypoint \emph{confidence} randomly, while in the ScanNet, the original still \xuelun{exhibits the best performance}.

\subsection{Keypoint confidence in matching}
Based on the experiment results \xuelun{obtained herein}, we conclude that the $\mathbf{p}$ and $\mathbf{d}$ dominate the matching performance of SuperGlue.
\xuelun{With regards to} $\mathbf{c}$, we cannot say that $\mathbf{c}$ is insignificant for matching; however, currently, in the SuperGlue architecture, it does not \xuelun{make a significant contribution} to matching.
\sxl{
\xuelun{In order to save} time in processing confidence, the confidence score can be omitted.
}

It is intuitive that the \emph{confidence} of the keypoints is initially used for keypoint selection.
In the matching stage, the keypoints are selected, all \xuelun{of which} are treated equally for the matching method or network; therefore, the keypoint confidence is not important.

However, \xuelun{situations may arise} in which SuperGlue does not use the information in $\mathbf{c}$ appropriately.
In these cases, a better approach could be designed to extract valuable information from $\mathbf{c}$ by combining the confidence value with the final matching results for a joint consideration and calculation.
In this work, we chose to remove $\mathbf{c}$ when matching keypoints \xuelun{in order} to achieve the purpose of detector-oblivious \xuelun{processing}.

\vspace{-1mm}
\section{Experiments for the multi-arm network}

This section \xuelun{compares} our MAN against state-of-the-art techniques in relative pose estimation and visual localization tasks.
We trained the indoor model on the ScanNet~\cite{SCAN2017} dataset and the outdoor model on the MegaDepth~\cite{MEGA2018} following the details provided by SuperGlue~\cite{GLUE2020}.
\sxl{
In details, the keypoint detector we used in training is the SuperPoint~\cite{SPOINT2018}.
}

\subsection{Relative pose estimation}


\parsection{Baselines}
Following SuperGlue~\cite{GLUE2020}, we compare our multi-arm detector-oblivious network against state-of-the-art matchers applied to SIFT and SuperPoint features -- the Nearest Neighbor (NN) matcher and various outlier rejectors: the mutual NN constraint, PointCN~\cite{GOODC2018}, OANet~\cite{OANET2019} and ACNe~\cite{ACNE2020}.
For all methods \xuelun{considered}, we report the results using their official codes or models.



\parsection{Outdoor scenario}
As shown in Table~\ref{tab:MAN-OUT-IN}, both MegaDepth and YFCC, our approach outperforms all baselines at all relative pose thresholds and the matching precision, when applied to both SuperPoint and SIFT.
Compare with SuperGlue, after training with two auxiliary tasks; our multi-arm network performance \xuelun{shows steady improvement}.

\parsection{Indoor scenario}
As shown in Table~\ref{tab:MAN-OUT-IN}, our approach achieves the best pose estimation and matching precision in the two datasets when integrating with both SuperPoint and SIFT.
Compared with SuperGlue, after adding the two auxiliary tasks, both the pose estimation and the matching precision are improved.
SuperGlue obtains better matching scores \xuelun{indicating that} it finds more correspondences under the same keypoints.
Our method tends to find more reliable correspondences with auxiliary constraints, \xuelun{thereby} reducing the number of matches and resulting in lower matching scores than SuperGlue.

\parsection{Comparison with Detector-free methods}
\sxl{
We also compared our method against recent detector-free methods~\cite{DFM2021, LOFTR2021}, which aim to find correspondences without using a keypoint detector. This is considered an intermediate solution between \emph{sparse-matching} and \emph{dense-matching} and can be called ``\emph{semi-dense matching}.''
}

\sxl{
In terms of results, our method performs better than DFM~\cite{DFM2021}, likely because the transformation between the test image pairs is more complex than the homography transformation predicted by DFM in its first stage, so DFM does not achieve its best results. If DFM can predict more complex transformations in its first stage, it should achieve better results.
}

\sxl{
LoFTR~\cite{LOFTR2021} achieves better results than our method, \xuelun{potentially due to} its coarse-to-fine architecture. In the coarse-level stage, LoFTR gets more dense coarse correspondences compared to detector-based methods and then refines them in the fine-level stage to obtain a better pose estimate.
}

\subsection{Visual localization}

\parsection{Datasets}
The Aachen Day-Night dataset~\cite{VLOCAL2012,VLOCAL2018} evaluates local feature matching for day-night localization.
For each of the 98 night-time images contained in the dataset, up to 20 relevant day-time images with known camera poses are given.
Following SuperGlue~\cite{GLUE2020}, we extracted up to 4096 keypoints per image with SuperPoint, matched them with \shortname, triangulated an SfM model from posed day-time database images, and register night-time query images with the image matches and COLMAP~\cite{SFM2016}.

\parsection{Metrics}
Following SuperGlue~\cite{GLUE2020}, we used the code and evaluation protocol from~\cite{VLOCAL2018} and report the percentage of night-time queries localized within given error bounds on the estimated camera position and orientation.
\sxl{
In detail, the Aachen Day-Night benchmark measures the percentage of query images localized within $X$m and $Y$\degree of their ground truth pose.
This benchmark defines three pose accuracy intervals by varying the thresholds: \emph{High-precision} (0.5m, 2\degree), \emph{medium-precision}(1m, 5\degree), and \emph{coarse-precision}(5m, 10\degree).
}

\parsection{Baselines}
Following SuperGlue~\cite{GLUE2020}, we compare our method against SuperPoint~\cite{SPOINT2018}, D2-Net~\cite{D2NET2019}, R2D2~\cite{R2D22019}, UR2KID~\cite{UR2KID2020} and SuperGlue~\cite{GLUE2020}.

\parsection{Results}
As reported in Table~\ref{tab:VL}, when compared with SuperGlue~\cite{GLUE2020}, our method achieves a better result than it under the $0.5m/2\degree$ threshold and achieves comparable results under the $1m/5\degree$ and $5m/10\degree$ thresholds.
\sxl{
We conjecture that the multi-arm architecture we propose can improve the performance of the network at high-precision ($0.5m/2\degree$), but not at medium-precision ($1m/5\degree$).
}
Compared with other methods, our method performs similarly or better despite using significantly fewer keypoints.
\sxl{
In general, our method generalizes well in visual localization.
}
\begin{table}
\centering
\renewcommand{\arraystretch}{1.3} 
\caption{
    Evaluation for visual localization.
}
\resizebox{\linewidth}{!}{\begin{tabular}{lccccc}
    \toprule[1.5pt]
    \multirow{2}{*}[-.4em]{Method}          & \multicolumn{3}{c}{Correctly localized queries (\%)}  & \multirow{2}{*}[-.4em]{\# features}   \\
    \cmidrule(lr){2-4}
                                            & 0.5m/2\degree & 1m/5\degree & 5m/10\degree                                                    \\
    \midrule
    SuperPoint~\cite{SPOINT2018}            & 43.9      & 59.2      & \UL{76.5}                     & 20k                                   \\
    D2-Net~\cite{D2NET2019}                 & 45.9      & 68.4      & \BF{88.8}                     & 15k                                   \\
    R2D2~\cite{R2D22019}                    & \BF{46.9} & 66.3      & \BF{88.8}                     & 15k                                   \\
    UR2KID~\cite{UR2KID2020}                & \BF{46.9} & 67.3      & \BF{88.8}                     & 4k                                    \\
    SuperPoint + SuperGlue~\cite{GLUE2020}  & 45.9      & \BF{70.4} & \BF{88.8}                     & 4k                                    \\
    \midrule
    SuperPoint + MAN                        & \UL{46.6} & \UL{69.8} & \BF{88.8}                     & 4k                                    \\
    \bottomrule[1.5pt]
\end{tabular}}
\label{tab:VL}
\vspace{-\baselineskip}
\end{table}

\section{Ablation study}
\label{exp:abla}

As shown in Table~\ref{tab:AB}, we trained eight models for quantitative comparison.
Model \textbf{\#4} is the finalized one, which compares other baseline methods in experiments.
We tested these eight models in both outdoor and indoor scenarios \xuelun{in order} to evaluate our design decisions.

\begin{table*}
\centering
\renewcommand{\arraystretch}{1.3} 
\caption{
\sxl{
    Ablation study of the multi-arm network.
    The first column represents the number of the models.
    The OVE and DEE refer to overlap estimation and depth region estimation, the two auxiliary tasks proposed in this work.
    The Multi-Arm represents the model that finds correspondences by using multiple feature branches.
}
}
\resizebox{0.95\linewidth}{!}{\begin{tabular}{c|ccc|rrrrr|rrrrr|c}
    \toprule[1.5pt]
    \multirow{2}{*}{\# Model}   & \multirow{2}{*}{OVE}  & \multirow{2}{*}{DEE}  & \multirow{2}{*}{Multi-Arm}    & \multicolumn{5}{c|}{MegaDepth}                                                                                            & \multicolumn{5}{c|}{ScanNet}                                                                                          & \multirow{2}{*}{Time (ms)}    \\
    \cmidrule(l){5-9}
    \cmidrule(l){10-14}
                                &                       &                       &                               & @$5^\circ$        & @$10^\circ$       & @$20^\circ$       & P                 & M                                         & @$5^\circ$        & @$10^\circ$       & @$20^\circ$       & P                 & M                                     &                               \\
    \midrule
    1                           & -                     & -                     & -                             & 39.90 	        & 52.49 	        & 63.07 	        & 92.88 	        & \multicolumn{1}{r|}{25.91} 	            & 16.45 	        & 33.62 	        & 51.96 	        & 84.13 	        & \multicolumn{1}{r|}{\BF{31.68}}       & 143\,\,\,/\,\,\,\,\,\,66      \\
    2                           & $\surd$               & -                     & -                             & \UL{41.10}        & \UL{53.29}        & \UL{64.06}        & \UL{93.87}        & \multicolumn{1}{r|}{\UL{27.12}}           & \BF{17.90}        & \BF{36.73}        & \BF{56.62}        & \BF{85.36}        & \multicolumn{1}{r|}{\UL{29.94}}       & 143\,\,\,/\,\,\,\,\,\,66      \\
    3                           & $\surd$               & $\surd_{\text{(2 valid regions)}}$         & -                             & 35.90 	        & 45.17 	        & 51.34 	        & 77.81 	        & \multicolumn{1}{r|}{20.55} 	            & 12.00 	        & 24.03 	        & 38.88 	        & 82.99 	        & \multicolumn{1}{r|}{30.03}            & 143\,\,\,/\,\,\,\,\,\,66      \\
    4                           & $\surd$               & $\surd_{\text{(2 valid regions)}}$         & $\surd$                       & \BF{42.55}        & \BF{54.65}        & \BF{64.56}        & \BF{94.54}        & \multicolumn{1}{r|}{\BF{27.23}}           & \UL{17.65}        & \UL{35.09}        & \UL{55.61}        & \UL{84.73}        & \multicolumn{1}{r|}{29.11}            & 221\,\,\,/\,\,\,103           \\
    5                           & -                     & -                     & $\surd$                       & 38.63 	        & 50.54 	        & 60.04 	        & 90.58 	        & \multicolumn{1}{r|}{24.04} 	            & 15.51 	        & 32.96 	        & 49.67 	        & 83.61 	        & \multicolumn{1}{r|}{29.05}            & 221\,\,\,/\,\,\,103           \\
    \sxl{6}                           & \sxl{-}                     & \sxl{$\surd_{\text{(2 valid regions)}}$}            & \sxl{-}                       		& \sxl{40.37} 	        & \sxl{52.81} 	        & \sxl{63.49} 	        & \sxl{93.35} 	        & \multicolumn{1}{r|}{\sxl{26.34}} 	            & \sxl{16.77} 	        & \sxl{34.24} 	        & \sxl{52.64} 	        & \sxl{84.59} 	        & \multicolumn{1}{r|}{\sxl{29.19}}            & \sxl{143\,\,\,/\,\,\,\,\,\,66}      \\
    \sxl{7}                           & \sxl{-}                     & \sxl{$\surd_{\text{(3 valid regions)}}$}            & \sxl{-}                       		& \sxl{31.37} 	        & \sxl{40.34} 	        & \sxl{49.43} 	        & \sxl{73.52} 	        & \multicolumn{1}{r|}{\sxl{19.46}} 	            & \sxl{12.84} 	        & \sxl{25.30} 	        & \sxl{36.96} 	        & \sxl{65.78} 	        & \multicolumn{1}{r|}{\sxl{23.21}}            & \sxl{143\,\,\,/\,\,\,\,\,\,66}      \\
    \sxl{8}                           & \sxl{-}                     & \sxl{$\surd_{\text{(4 valid regions)}}$}            & \sxl{-}                       		& \sxl{22.77} 	        & \sxl{30.08} 	        & \sxl{34.52} 	        & \sxl{49.01} 	        & \multicolumn{1}{r|}{\sxl{15.57}} 	            & \sxl{9.15}  	        & \sxl{20.32} 	        & \sxl{28.25} 	        & \sxl{43.21} 	        & \multicolumn{1}{r|}{\sxl{16.57}}            & \sxl{143\,\,\,/\,\,\,\,\,\,66}      \\
    \bottomrule[1.5pt]
\end{tabular}}
\label{tab:AB}

\end{table*}

\begin{table}
\centering
\renewcommand{\arraystretch}{1.3} 
\caption{
\sxl{
    Ablation experiments of the model trained and tested with different detectors.
    The \textbf{SP} represents SuperPoint~\cite{SPOINT2018} detector.
}
}
\resizebox{0.8\linewidth}{!}{\begin{tabular}{cc|lllll}
    \toprule[1.5pt]
    \multirow{2}{*}{\sxl{Train}}			& \multirow{2}{*}{\sxl{Test}}		& \multicolumn{5}{c}{\sxl{MegaDepth}}                                                                    \\
    \cmidrule(lr){3-7}
                                    &                         	& \sxl{@$5^\circ$}        & \sxl{@$10^\circ$}       & \sxl{@$20^\circ$}       & \sxl{P}                 & \sxl{M}                 \\
    \midrule
    \sxl{SIFT}                            & \sxl{SIFT}                      & \sxl{$27.65$}           & \sxl{$38.90$}           & \sxl{$54.93$}           & \sxl{$95.25$}			& \sxl{$17.61$}           \\
    \sxl{SIFT}                            & \sxl{SP}                    	& \sxl{$40.88$}           & \sxl{$53.15$}           & \sxl{$61.44$}           & \sxl{$93.01$}           & \sxl{$26.58$}  			\\
    \sxl{SP}                              & \sxl{SIFT}                      & \sxl{$26.01$}        	& \sxl{$36.25$}		    & \sxl{$50.63$}		    & \sxl{$94.78$}		    & \sxl{$16.65$}		    \\
    \sxl{SP}                              & \sxl{SP}                      	& \sxl{$42.55$}        	& \sxl{$54.65$}		    & \sxl{$64.56$}		    & \sxl{$94.54$}		    & \sxl{$27.23$}		    \\
    \bottomrule[1.5pt]
\end{tabular}}
\label{tab:oblivious}
\end{table}

Comparing models \textbf{\#1} and \textbf{\#2} shows that the overlap estimation task proposed \xuelun{herein can} improve the matching precision and leads to estimate more accurate poses between images.
Model \textbf{\#3} is trained using two auxiliary tasks with Single-Arm architecture.
\xuelun{We showed experimentally} that \xuelun{allowing} one feature \xuelun{to simultaneously} learn the two auxiliary tasks will lead to performance degradation, however, adding the Multi-Arm architecture and \xuelun{allowing} multiple features to learn different auxiliary tasks \xuelun{improves} the performance \xuelun{as shown in} \textbf{\#4}.
If only the Multi-Arm architecture is use and multiple features are not trained with auxiliary tasks, the performance will decline slightly as shown in model \textbf{\#5}.


\sxl{
In the depth estimation auxiliary task, we found that \xuelun{allowing} the network to estimate two depth regions (close-scene and far-scene regions) improves the matching results \xuelun{as shown in} model \textbf{\#6}.
We also attempted to allow the network to predict three and four regions in the auxiliary task; this resulted in a degradation of the performance \xuelun{as shown in} models \textbf{\#7} and \textbf{\#8}.
We propose that this is because too complex auxiliary tasks affect the learning of the primary task.
}

When designing auxiliary networks, \xuelun{it is important to ensure that they} play an auxiliary role and we expect their optimization to work toward the main matching task, not only auxiliary tasks.
Therefore, we used only three layers of MLP to design the auxiliary network.
\xuelun{It is difficult for} such a light-weight auxiliary network to learn complex capabilities, \xuelun{such as accurately estimating} the overlap area or depth region \xuelun{in which} the keypoints fall.
However, we think it is interesting to find the keypoints in the overlap area first or predict the depth region of the keypoints accurately.
If the network can exclude keypoints outside the overlap area, we could eliminate many potential mismatches in advance.
If the network can estimate the keypoint depth region precisely, we could only match the corresponding area's keypoints, which could also prevent matches in the non-corresponding regions.

\parsection{Inference time}
We measured the run-time of our approach in outdoor and indoor test scenarios.
In both scenarios, the measurements were performed using an NVIDIA GeForce RTX 3090 Graphics Card.
In the outdoor scenario, the measurement was performed across $868$ images with $2048$ keypoints per image.
\xuelun{The model with} Single-Arm (similar to SuperGlue) runs at $143~ms$ per image, \xuelun{whereas} the model with Multi-Arm runs at $221~ms$ per image.
In the indoor scenario, the measurement was performed across $1500$ images with $1024$ keypoints per image. \xuelun{The model with} Single-Arm runs at $66~ms$ per image, and the model with Multi-Arm runs at $103~ms$ per image.
The Multi-Arm network brings performance improvements in outdoor scenarios; it also brings more computing time.
However, the proposed OVE could improve the matching performance for the network indoors and outdoors without increasing inference time.

\begin{figure*}
\centering
\newcommand{\wid}{\linewidth}
\includegraphics[width=\wid]{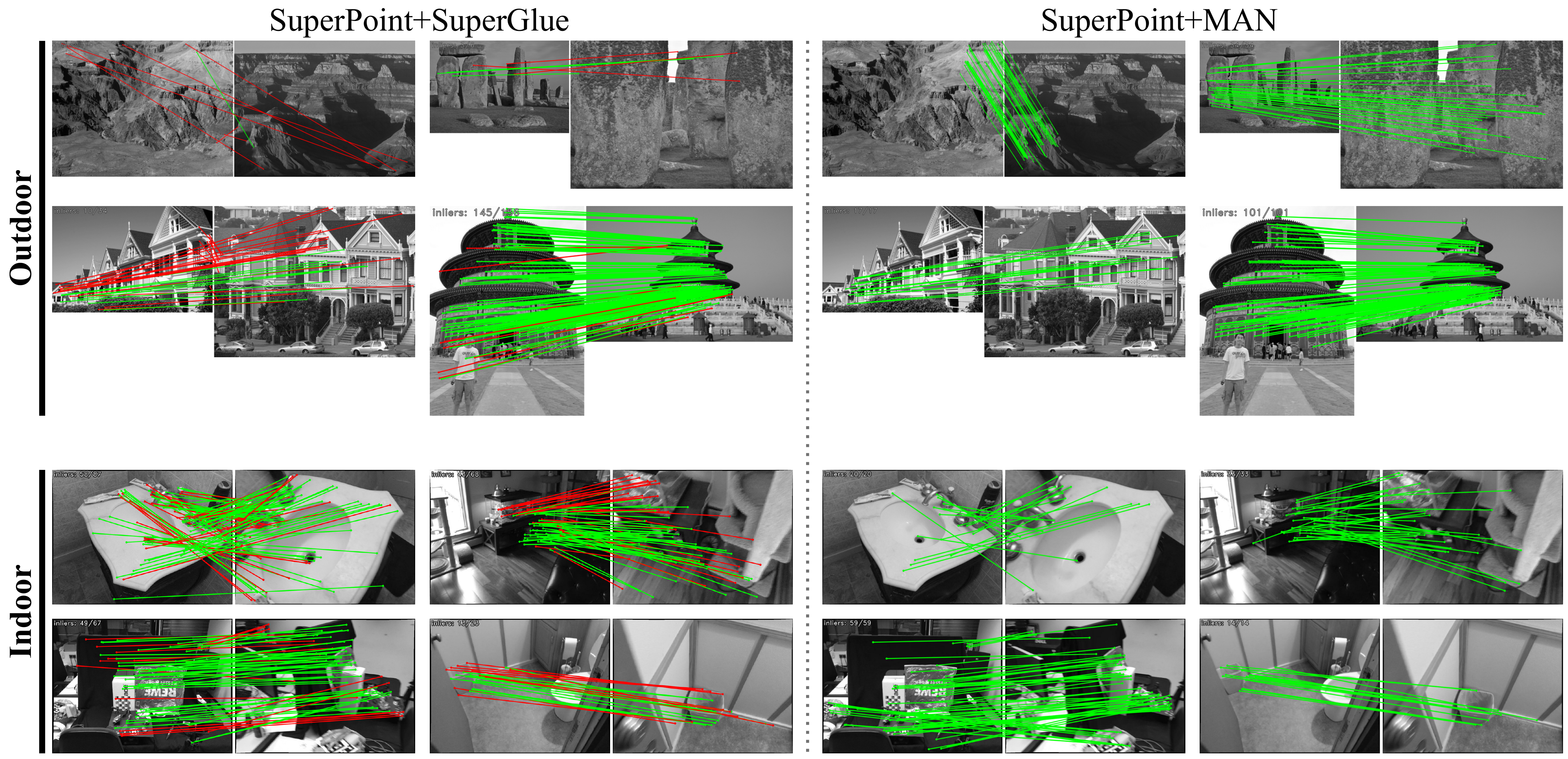}\vspace{0mm}
\vspace{-1mm}\caption{
    \BF{Qualitative image matches.}
    We compared \shortname and SuperGlue~\cite{GLUE2020} in indoor and outdoor environments.
    {\color{red}Red} and {\color{green}green} lines indicate incorrect and correct matches, respectively.
}
\label{fig:Show}
\vspace{-\baselineskip}
\end{figure*}

\parsection{Visualization}
We present some qualitative image matching results in Fig.~\ref{fig:Show}.
We compare \shortname to SuperGlue~\cite{GLUE2020} in indoor and outdoor environments.
\shortname consistently estimates more correct matches, successfully coping with the low-overlap situation and dramatic viewpoint changes.

\parsection{Detector-oblivious}
\sxl{
To demonstrate "detector-oblivious", we conducted ablation experiments of the model trained and tested with different detectors.
As shown in Table~\ref{tab:oblivious}, our method combined with SuperPoint presents better performance than combined with SIFT in testing.
Compared to training and testing on the same detector, there is no significant performance degradation when it is trained and tested on different detectors.
The experiment results demonstrate our method does not need to be retrained when a new detector is adopted in testing.
}

\section{Conclusions}
This work proposes a multi-arm network to extract information from overlap and depth regions in images, which provides extra supervision signals for training.
Solving the auxiliary tasks of OVerlap Estimation (OVE) and DEpth region Estimation (DEE) help significantly improve the keypoint matching performance while bringing the little computational cost in inference.
Besides, we designed a detector-oblivious description network and integrated it into the matching pipeline.
This design makes our network can directly adapt to unlimited keypoint detectors without a time-consuming retraining process.

\bibliographystyle{A-packages/IEEEtran}
\bibliography{A-packages/egbib}

\end{document}